# A Novel Multi-clustering Method for Hierarchical Clusterings, Based on Boosting

Elaheh Rashedi*, Abdolreza Mirzaei**
*Isfahan University of Technology, e.rashedi@ec.iut.ac.ir
** Isfahan University of Technology, mirzaei@cc.iut.ac.ir

***Abstract***: Bagging and boosting are proved to be the best methods of building multiple classifiers in classification combination problems. In the area of "flat clustering" problems, it is also recognized that multi-clustering methods based on boosting provide clusterings of an improved quality. In this paper, we introduce a novel multi-clustering method for "hierarchical clusterings" based on boosting theory, which creates a more stable hierarchical clustering of a dataset. The proposed algorithm includes a boosting iteration in which a bootstrap of samples is created by weighted random sampling of elements from the original dataset. A hierarchical clustering algorithm is then applied on selected subsample to build a dendrogram which describes the hierarchy. Finally, dissimilarity description matrices of multiple dendrogram results are combined to a consensus one, using a hierarchical-clustering-combination approach. Experiments on real popular datasets show that boosted method provides superior quality solutions compared to standard hierarchical clustering methods.

## 1. Introduction

Using multi-learner systems is the main idea of ensemble learning, which creates multiple learners of the ensemble in order to combine their predictions. according to the recent knowledge, the more recent powerful general ensemble methods are bagging (bootstrap aggregating) and boosting [1-3].

Bagging is a multiple learner combination method which applies individual copies of a weak learner algorithm on bootstrap replications of samples. Sampling is based on random selection with replacement in which the samples' weight are distributed uniformly [4]. Boosting is also defined as "a general problem of converting a weak learning algorithm into one with higher accuracy" [5]. The main difference is use of reweighting. An adaptive boosting algorithm, AdaBoost, is proposed in [6] which is applicable to a general class of learning problems. AdaBoost has two main phases, 1) iteratively generation of new hypothesis on a bootstrap of samples using a new distribution of training examples, and 2) combining hypothesis through majority vote. Boosting concepts can be applied to either supervised learning (classification) and unsupervised learning (clustering) problems, as it is proved to be the most powerful method of creating classifier ensemble.

There are many algorithms proposed in the area of classification ensemble problems which are based on bagging and boosting [1, 2, 4, 7-9]. There are also some multi-clustering algorithms base on bagging and boosting which are introduced on the flat clusterings. These methods commonly used k-means as a partitional (non hierarchical) clustering algorithm, and the combination is done through voting [2, 3, 10-13].

To the best of our knowledge, few hierarchical cluster combination methods are invented and none of them addressed the boosted design of weak hierarchical clusterers [14-16]. Multi-hierarchical-clustering methods has more challenges than flat ones, as it is harder to verify how well a data point is clustered in hierarchy, which cause the sample reweighting to be more notable.

In this paper a multi-hierarchical-clustering algorithms is proposed based on boosting theory. This algorithm iteratively generates new hierarchical clusterings on a bootstrap of samples. Subsampling is done using a new updated distribution of training examples. Experiments show that the boosted method provides superior quality solutions compare to standard hierarchical clustering methods.

The paper organization is as follow: In Section 2, a new boosted multi-clustering algorithm is presented for hierarchical clusterings. Section 3 shows the experimental results of applying the proposed method on some real popular datasets and Comparatives results are also illustrated. Finally, the concluding remarks are presented in Section 4.

## 2. The Boosted Hierarchical Cluster Ensemble Method

In this paper we propose a new clustering ensemble method for hierarchical clusterings. The method is based on boosting, which iteratively select a new training set using weighted random sampling and provide multiple hierarchical clusterings which results to a final aggregated clustering. This is a general method which can apply any hierarchical clustering algorithm to generate ensemble and any hierarchical combination method can be used to aggregate the individual clusterings from the ensemble. At the first iteration of boosting, a new training set is provided by random sampling from the original dataset. A hierarchical clustering algorithm is applied on the selected samples to create the first hierarchical partition of data. For the next other iterations, the samples' weights are updated according to the boosting efficacy of the previous hierarchical partitioning of data, and the next base clustering is generated relating to these new weights. Final clustering solution is produced by combining all hierarchical partitions in hand. In combination stage, dissimilarity description matrices of multiple clustering results are combined into one description matrix. The combination is done based on the Rényi divergences entropy approach. The boosted combination algorithm is as below.

## 2.1 Boosted Hierarchical Clustering Ensemble Algorithm

*HBoosting*: A data set of $N$ samples, $x_1, x_2, \ldots, x_N$, shown as $D$, is Given. The Output is a consensus hierarchical clustering $H^*$.

1. Initialization:
   - $i = 1$: iteration number
   - $T$: maximum number of iteration
   - $w_n^0 = 1/N$: initial weight of each sample for $1 \leq n \leq N$
2. Iterative process done for $T$ times
   - Bootstrapping
     Weighted random sampling from the original dataset $D$ according to the probability of ($w_n^{i-1}/\sum_{n=1}^N w_n^i$) for every instance
   - Hierarchical partitioning
     Generate the $i$'th base hierarchical clustering of the ensemble, $H_i$, on selected subsample
   - Aggregating
     Combination of all individual hierarchical clusterings
     $H_i^{agg} = combine\ (H_0, \ldots, H_i)$
   - Additive updating of weights
     give higher weight to instances of lower boosted value, to have higher selection probability. $BV_n^i$ is a boosted value used to evaluate the quality of clustering of the **n**'th instance in the hierarchy
     $w_n^i = w_n^{i-1} + (-BV_n^i)\quad 1 \leq n \leq N$
   - Go to step 2
     $i = i + 1$
3. Obtaining final consensus hierarchical clustering
   $H^* = H_T^{agg}$

The boosted value of each sample in the **i**'th iteration, $BV_n^i$, is a measurement index to show how well a data point is clustered in the **i**'th hierarchy. The suggested is calculated as the correlation between the Euclidian distances of each sample in original dataset, $D$, and a defined distance of the sample in aggregated hierarchical dendrogram $H_i^{agg}$.

## 3. Experimental Results

The Boosted hierarchical cluster ensemble method has been evaluated on various benchmark datasets given in Table 1. Datasets are collected from three popular real dataset repository, Gunnar Raetsch's Benchmark Datasets [17], Univ. Calif. Irvine Repository of Machine Learning Databases [18] and Real Medical Datasets [19]. The trial datasets are of different sample sizes, from 24 to 692. The source and the characteristics of experimented datasets are represented in Table 1.

TABLE I: Source and Characteristics of Datasets Used in this Experiment

| Data set | #dataset | #instances | #features | #class | source |
|---|---|---|---|---|---|
| Breast_cancer | 1 | 263 | 9 | 2 | [17] |
| contraction | 2 | 98 | 27 | 2 | [19] |
| Flare_solar | 3 | 144 | 9 | 2 | [17] |
| Laryngeal1 | 4 | 213 | 16 | 2 | [19] |
| Laryngeal2 | 5 | 692 | 16 | 2 | [19] |
| Laryngeal3 | 6 | 353 | 16 | 3 | [19] |
| Titanic | 7 | 24 | 3 | 2 | [17] |
| Wine | 8 | 178 | 13 | 3 | [18] |

At the starting point of the experiment, a base hierarchical clustering algorithm is needed to create dendrogram ensembles. Some popular agglomerative hierarchical clustering methods are *Centroid*, *Single*, *Average*, *Complete*, *Weighted*, *Median* and *Ward*. The proposed method is experimented under all of these Clusterer types to be evaluated by the best one.

The clustering algorithms are to be applied on a subsample of data, instead of the whole. Here, the bootstrap subsample is set to be 20% of the original dataset samples. Creating the dendrograms on the bootstrap subsample, the next step is to represent the dendrograms in form of description matrices. In this experiment the cophenetic difference (CD) similarity matrices are used.

The combination method is then applied to base similarity descriptor matrices. The combination is based on the *Rényi divergences entropy approach* with parameter $\beta$ [20]. Setting parameter $\beta$ to $\{-\infty, 1, +\infty\}$, the combination function is converted to common operators *Minimum*, *Average* and *Maximum*. In this experiment the method is evaluated by each of these three combination types.

These combination methods are applied to similarity matrices. The consensus matrix is then put into a dendrogram recovery function, to retrieve the final hierarchical clustering result. The tested Recovery methods are *Average*, *Single*, *Complete*, *Ward*, *Centriod* and *Median*.

According to the mentioned parameters, Clusterer type, combination type and Recovery method, the experiments are conducted on $7 * 3 * 6 = 126$ different parameter values to validate the accuracy of consensus hierarchical partition results. So, the ensemble method is applied **126** times on each dataset. Number of boosting iterations is set to **200**. Completing all boosting iterations, the quality of final consensus hierarchical clustering is compared to standard hierarchical clusterings. The quality measurement method and comparative results are explained in the following paragraph.

In order to evaluate the proposed method, a quality measure is needed to verify whether the consensus hierarchical clustering structure fits the original data. The best known measure is cophenetic correlation coefficient ($CPCC$). In this experiment the $CPCC$ is computed as the correlation between the Euclidian distance on original data and the cophenetic difference (CD) distances on hierarchy. The $CPCC$ interval value is $[-1,1]$, where upper values show the better agreement between two tested matrices. The $CPCC$ is measured to show the quality of experimental results.

In order to evaluate each parameter of the proposed method, a multiple range test, is performed on $CPCC$ value of all **126** results of each dataset in hand.

Duncan is a multiple range test that can be used to determine the significant differences between group mean [21], means within the same group are not significantly different and, those from different groups are significantly different at an assumed level of $\alpha = 0.05$. Groups are sorted in descending order, i.e. maximum values are in group A.

In this experiment, Duncan test is used to compare results of discussed parameters, Clusterer types, combination types and Recovery methods. The test is as follow. For each Clusterer type, the ensemble method is applied $3 * 6 = 18$ times on each dataset. The mean $CPCC$ values of each Clusterer type are calculated and put in descending order sorted groups. The frequency of how many times a parameter takes place in a group is counted for all datasets. The frequency of each Clusterer type in each group is shown in Fig. 1. Similarly, Fig. 2 and 3 shows the frequency of each method in each group, for combination types and Recovery methods.

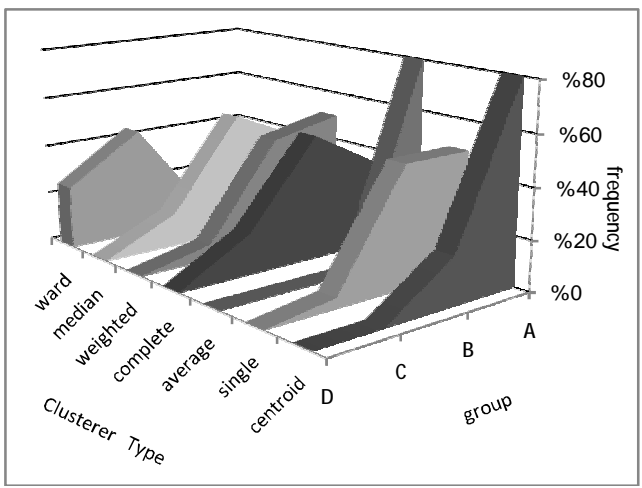

Figure 1. Frequency of times clusterer types take place in groups

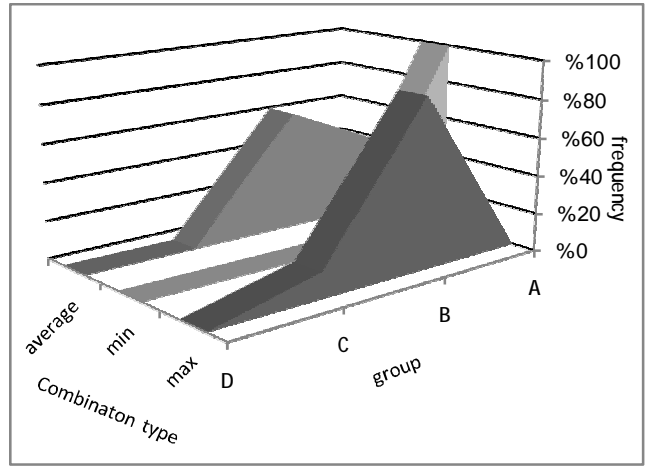

Figure 2. Frequency of times combination types take place in groups

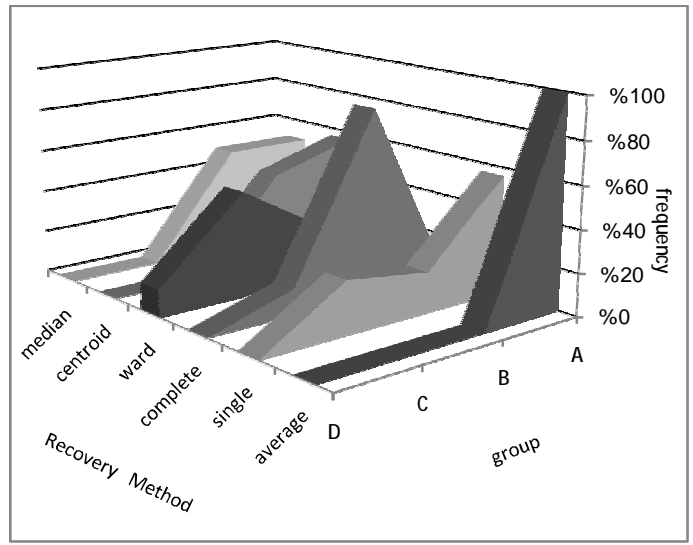

Figure 3. Frequency of times recovery methods take place in groups

It is observesd from Fig. 1, 2 and 3 that the aggregation of centriod or average clusterer types, min combination type and average recovery method commonly generate results of better quality. A comparison of these results with quality value of average and centroid basic hierarchical clustering methods is shown in Table 2. The maximum $CPCC$ values obtained from the proposed method and base methods are also compared.

Comparing the maximum values shows quality improvement in **6** situations from **8** (i.e. all datasets but the datasets number 3 and 7), with significant difference level of **0.05**.

Although the experiments show that the quality of preferable selected methods of the HBoosting algorithm (i.e. *Average/Min/Average* and *Centroid/Min/Average*) and the standard *Average* and *Centroid* linkage methods are close, overall, the *HBoosting* algorithm leads to the best partitioning quality in almost all cases.

TABLE II: Comparison between common desired solutions and basic hierarchical clustering approach, Average and Centroir with no subsampling

| Dataset | | 1 | 2 | 3 | 4 | 5 | 6 | 7 | 8 |
|---|---|---|---|---|---|---|---|---|---|
| HBoosting | average/min/average | 0.748 | 0.792 | 0.891 | 0.918 | 0.887 | 0.917 | 0.818 | 0.736 |
| | centriod/min/average | 0.741 | 0.794 | 0.889 | 0.926 | 0.900 | 0.923 | 0.817 | 0.767 |
| | Max of all methods | **0.748** | **0.802** | **0.891** | **0.927** | **0.900** | **0.923** | **0.818** | **0.767** |
| Single methods | average | 0.716 | 0.779 | 0.889 | 0.896 | 0.870 | 0.889 | 0.818 | 0.759 |
| | centriod | 0.715 | 0.756 | 0.887 | 0.892 | 0.875 | 0.885 | 0.778 | 0.755 |
| | Max of all methods | **0.716** | **0.779** | **0.889** | **0.896** | **0.870** | **0.889** | **0.818** | **0.759** |

Table 3 contains the frequency of times which a method creates a hierarchy of the best accuracy. The preferred clusterer type, i.e. average and centroid get better accuracies in **80** percent cases, as it is put in group "A" in **80** percent datasets. Respectively, the min combination type and average recovery method get better accuracies in **100** percent cases, as they are put in group "A" in **100** percent datasets. It is indicated that the HBoosting algorithm leads to the best partitioning quality in almost all cases, though the single clustering algorithms create better results in **75** percent cases. This comparison shows that the proposed method is more stable in practice.

TABLE III: Frequency of Times which the HBoosting Method and the Single Methods Creates a Hierarchy of the Best Accuracy

| Preferable Methods | | Frequency |
|---|---|---|
| HBoosting | Clusterer type = Average\|Centroid | 80% |
| | Comb type = Min | 100% |
| | Recovery method = Average | 100% |
| Single methods | Average link | 75% |
| | Centroid link | 75% |

## 4. Conclusions

In this paper a novel multi-hierarchical-cluster method is proposed based on boosting theory. In the boosting step of the algorithm, we introduce a new validation procedure to evaluate how well an individual data point has been clustered in the hierarchy. The calculated value is used in reweighting samples. Completing all boosting iterations, an aggregation of all created clusterings forms the final clustering result.

To the best of our knowledge, numerous ensemble methods are exists which construct a set of classifiers or flat clusterings based on boosting, while ignoring the situations in which a hierarchy of clusters is needed. So the proposed algorithms can acquire the needs. In the other hand, the introduced algorithm can deal with large volume datasets as it used a bootstrap of samples instead of whole. It should mention that large training datasets are difficult to create a single clusterer on. Moreover of being used instead of single clusterers, the method is more stable than single ones and also gains a better quality.

We use several real datasets to show that boosting is a good method to build multiple hierarchal clusterings. Comparing the results of an ensemble with basic single hierarchical clusterings prove the quality improvement and more stable clustering creation.

## References

[1] Y. M. Sun, Y. Wang, and A. K. C. Wong, "Boosting an associative classifier," *IEEE Transactions on Knowledge and Data Engineering,* vol. 18, pp. 988-992, 2006.
[2] Y. Freund and R. E. Schapire, "Experiments with a new boosting algorithm," in *Thirteenth International Conference on Machine Learning*, Bari, Italy, 1996, pp. 148-156.
[3] T. G. Dietterich, "Ensemble Methods in Machine Learning," in *First International Workshop on Multiple Classifier Systems*, 2000, pp. 1-15.
[4] L. Breiman, "Bagging predictors," *Machine Learning,* vol. 24, pp. 123-140, 1996.
[5] L. I. Kuncheva, *Combining Pattern Classifiers: Methods and Algorithms*: Wiley-Interscience 2004.
[6] Y. Freund and R. E. Schapire, "A decision-theoretic generalization of on-line learning and an application to boosting," *Journal of Computer and System Sciences,* vol. 55, pp. 119-139, 1997.
[7] S. Dudoit and J. Fridlyand, "Bagging to improve the accuracy of a clustering procedure," *Bioinformatics,* vol. 19, pp. 1090-1099, 2003.
[8] Y. Freund and R. E. Schapire, "A short introduction to boosting," *Journal of Japanese Society for Artificial Intelligence,* vol. 14, pp. 771-780, 1999.
[9] J. R. Quinlan, "Bagging, Boosting, and C4.5," in *Thirteenth National Conference on Artificial Intelligence* 1996.
[10] B. Minaei-bidgoli, E. Topchy, and W. F. Punch, "Ensembles of Partitions via Data Resampling," in *International Conference on Information Technology* 2004, p. 188.
[11] J. Chang and D. M. Blei, "Mixtures of Clusterings by Boosting," in *Learning Worksho* Hilton Clearwater, 2009.
[12] A. Topchy, B. Minaei, A. K. Jain, and W. Punch, " Adaptive clustering ensembles," in *Internat. Conf. on Pattern Recognition*, 2004, pp. 272–275.
[13] M. Al-Razgan and C. Domeniconi, "Weighted clustering ensembles," in *SIAM International Conference on Data Mining*, 2006, pp. 258–269.
[14] A. Mirzaei and M. Rahmati, "Combining Hierarchical Clusterings Using Min-transitive Closure," in *19th International Conference on Pattern Recognition (ICPR 2008)* Tampa, Florida, USA: IEEE, 2008.
[15] A. Mirzaei and M. Rahmati, "A Novel Hierarchical-Clustering-Combination Scheme Based on Fuzzy-Similarity Relations," *IEEE Transactions on Fuzzy Systems,* vol. 18, pp. 27-39, 2010.
[16] A. Mirzaei, M. Rahmati, and M. Ahmadi, "A new method for hierarchical clustering combination," *Intelligent Data Analysis,* vol. 12, pp. 549-571, 2008.
[17] "Gunnar Raetsch's Benchmark Datasets.", Available: http://users.rsise.anu.edu.au/~raetsch/data/index
[18] "Univ. Calif. Irvine Repository of Machine Learning Databases.", Available: http://www.ics.uci.edu/~mlearn/MLRepository.htm
[19] "Real Medical Datasets.", Available: http://www.informatics.bangor.ac.uk/~kuncheva/activities/real_data.htm
[20] A. Mirzaei, "Combining Hierarchical Clusterings With Emphasis On Retaining The Structural Contents Of The Base Clusterings," in *Computer Engineering & IT Department*. vol. Doctor of Philosophy in Computer Engineering Tehran: Amir-kabir University of Technology, 2009.
[21] D. B. Duncan, "Multiple range and multiple F tests," *Biometrics* vol. 11, pp. 1-42, 1995.